\documentclass[letterpaper, 10 pt, conference]{ieeeconf} 
\IEEEoverridecommandlockouts 
\usepackage[]{graphicx}
\usepackage{cite}
\usepackage{amsmath}
\usepackage{amssymb}
\usepackage{textcomp}
\usepackage{xcolor}
\usepackage{url}
\newcommand{\opename}[1]{\operatorname{\mathsf{#1}}}
\usepackage{multirow}

\title{\LARGE \bf
360\textdegree{} Depth Estimation from Multiple Fisheye Images \\ with Origami Crown Representation of Icosahedron
}

\author{Ren Komatsu$^{1}$, Hiromitsu Fujii$^{2}$, Yusuke Tamura$^{3}$, Atsushi Yamashita$^{1}$, and Hajime Asama$^{1}$
\thanks{*A part of this study is supported by the Nuclear Energy Science \& Technology and Human Resource Development Project (through concentrating wisdom) from the Japan Atomic Energy Agency / Collaborative Laboratories for Advanced Decommissioning Science, and Initiative on Promotion of Supercomputing for Young or Women Researchers, Supercomputing Division, Information Technology Center, The University of Tokyo.}
\thanks{$^{1}$R. Komatsu, A. Yamashita, and H. Asama are with the Department of Precision Engineering, Graduate School of Engineering, The University of Tokyo, Tokyo 113-8656, Japan (email: {\tt\small \{komatsu, yamashita, asama\}@robot.t.u-tokyo.ac.jp}).}%
\thanks{$^{2}$H. Fujii is with the Department of Advanced Robotics, Faculty of Advanced Engineering, Chiba Institute of Technology, Narashino 275-0016, Japan (email: {\tt\small hiromitsu.fujii@p.chibakoudai.jp}).}%
\thanks{$^{3}$Y. Tamura is with the Department of Robotics, Division of Mechanical Engineering, Tohoku University, Sendai 980-8579, Japan (email: {\tt\small y.tamura@srd.mech.tohoku.ac.jp}).}%
}
\begin{document}
\maketitle
\thispagestyle{empty}
\pagestyle{empty}
\begin{abstract}
In this study, we present a method for all-around depth estimation from multiple omnidirectional images for indoor environments. In particular, we focus on plane-sweeping stereo as the method for depth estimation from the images. 
We propose a new icosahedron-based representation and ConvNets for omnidirectional images, which we name ``CrownConv'' because the representation resembles a crown made of origami. CrownConv can be applied to both fisheye images and equirectangular images to extract features. 
Furthermore, we propose icosahedron-based spherical sweeping for generating the cost volume on an icosahedron from the extracted features. The cost volume is regularized using the three-dimensional CrownConv, and the final depth is obtained by depth regression from the cost volume. Our proposed method is robust to camera alignments by using the extrinsic camera parameters; therefore, it can achieve precise depth estimation even when the camera alignment differs from that in the training dataset. We evaluate the proposed model on synthetic datasets and demonstrate its effectiveness.
As our proposed method is computationally efficient, the depth is estimated from four fisheye images in less than a second using a laptop with a GPU. Therefore, it is suitable for real-world robotics applications.
Our source code is available at \url{https://github.com/matsuren/crownconv360depth}.

\end{abstract}

\section{INTRODUCTION}
The depth estimation of the surrounding environment of a vehicle is becoming increasingly important in the field of robotics and computer vision, as depth information is required for tasks such as autonomous navigation and object detection. The use of LiDAR is one approach to obtain depth information; however, RGB cameras are also commonly used for depth estimation owing to their low cost, light weight, and availability. In particular, fisheye cameras or omnidirectional cameras
are used to estimate the depth of the surroundings owing to their wide field of view (FoV)~\cite{heng2019, cui2019, won2019icra, won2019iccv}.

Plane-sweeping stereo, which has been studied for numerous years~\cite{oldpaper, oldpaper2, oldpaper3, sweeping, sweeping2, hane2014}, is one approach to depth estimation from multi-view images. In plane-sweeping stereo, multi-view images are projected onto virtual planes at several distances from the reference image plane to generate a cost volume. Thereafter, depth maps are estimated using this cost volume.
Recently, convolutional neural networks (ConvNets) have been applied to plane-sweeping stereo using perspective images, with remarkable results achieved~\cite{gcnet, psm,deepmvs,mvsnet,dpsnet,octdps}. 

\begin{figure}[tb]
    \centering
  \includegraphics[scale=0.82]{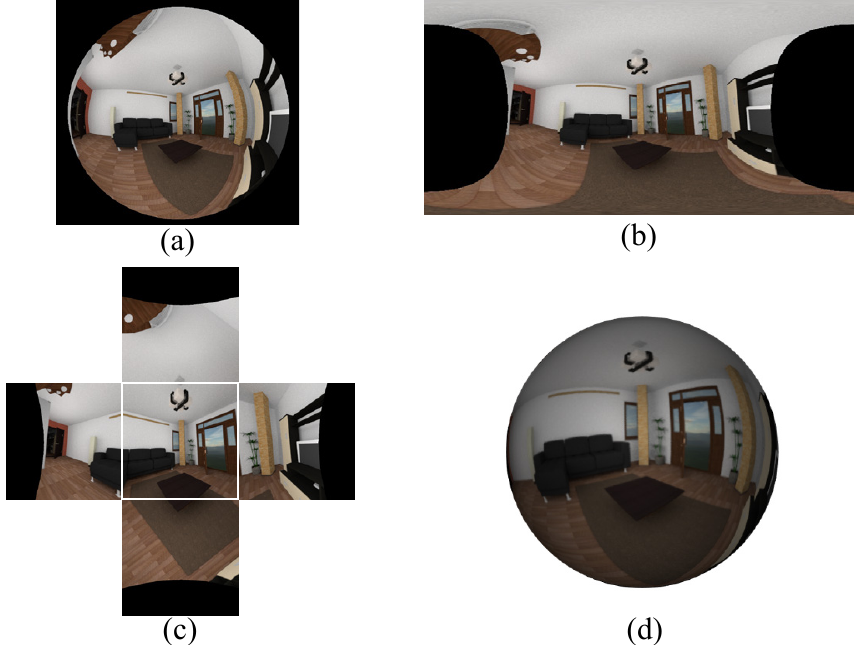}
    \caption{Example of fisheye image representation from OmniHouse dataset~\cite{won2019iccv}. (a) Original fisheye image with FoV of 220{\textdegree}. (b) Equirectangular image. (c) Cubemap representation. The back face is removed because all regions are beyond the camera FoV. (d) Fisheye image projected onto icosahedron at level 7. (a) and (d) appear similar; however, (d) indicates that the image is projected onto the icosahedron in 3D space.}
      \label{fig:fisheye_erp}
\end{figure}

As opposed to perspective images, the direct application of ConvNets to fisheye images is not desirable as equivariance to translation is not satisfied for fisheye images owing to distortion. Fig.~\ref{fig:fisheye_erp}(a) presents an example of fisheye images with a FoV of 220\textdegree. As can be observed from Fig.~\ref{fig:fisheye_erp}(a), straight lines in the three-dimensional (3D) world are captured as curved lines in the image because of distortion.
Several studies have applied ConvNets to fisheye images or equirectangular images directly, relying on the high flexibility of ConvNets to learn the distortion of the images~\cite{won2019icra, won2019iccv}. 
Won et al. proposed a method for estimating the all-around depth from four fisheye images~\cite{won2019icra}. 
They converted fisheye images into equirectangular images and applied ConvNets to extract the features directly. The cost volume was generated and refined using spherical sweeping~\cite{im2016}, followed by semi-global matching (SGM)~\cite{sgm}. Later, they proposed replacing SGM with 3D ConvNets~\cite{won2019iccv}.
However, such approaches are not robust to changes in the camera alignment, as demonstrated in this study. This is undesirable because every time the camera alignment is changed, additional training is required on datasets with the corresponding camera alignment.

In very recent years, ConvNets that were designed specifically for images other than perspective images have been studied intensively owing to the growing popularity of 360{\textdegree} cameras or omnidirectional cameras (for example, RICOH THETA and Insta360 ONE X). Certain studies~\cite{su2017, su2019} have focused on ConvNets that are applied to equirectangular images, which is a common image representation technique for omnidirectional cameras. Fig.~\ref{fig:fisheye_erp}(b) presents an example of equirectangular images. As can be observed in Fig.~\ref{fig:fisheye_erp}(b), equirectangular images exhibit the characteristic that stronger distortion exists when the area is closer to the top and bottom of the image. Su et al. proposed SphConv~\cite{su2017}, whereby different sizes of ConvNet kernels were applied to different rows to compensate for the distortion. 
However, the number of parameters was large because the weights were only shared in the row direction. To deal with this memory inefficiency, the authors proposed the Kernel Transformer Network~\cite{su2019} to share the same weights even in different rows. Although these approaches can handle distortion on equirectangular images, they are not computationally efficient because of oversampling in the top and bottom regions of the equirectangular images. Coors et al. proposed SphereNet~\cite{coors2018}, in which normal ConvNet kernels were distorted by sampling points on the tangent plane. Moreover, they proposed uniform sphere sampling to prevent oversampling for efficient computation. 

Cheng et al. used cubemap representations for omnidirectional images~\cite{cheng2018}. As can be observed in Fig.~\ref{fig:fisheye_erp}(c), cubemap represents an omnidirectional image as multiple perspective images by projecting the image onto six cube faces; therefore, normal ConvNets can be applied to these perspective images. However, this approach suffers from discontinuity between the faces, and ambiguity in the kernel orientations for the top and bottom faces.

Another approach involves projecting omnidirectional images onto a sphere surface ($\mathcal{S}^2$) and applying ConvNets in non-Euclidean space. The application of ConvNets to manifold or graph structures has been studied in the field of geometric deep learning~\cite{bronstein2017}.
Cohen et al. proposed the use of ConvNets in the frequency domain using a generalized fast Fourier transform~\cite{cohen2018}, whereas Esteves et al. proposed using ConvNets in the spherical harmonic domain~\cite{esteves2018}. These approaches extract features that are rotation invariant in the rotation group $\text{SO}\left(3\right)$; however, they require significant memory and computational costs.

Various other studies have used icosahedrons, because a subdivided icosahedron can be used to generate nearly uniformly distributed points on $\mathcal{S}^2$~\cite{lee2019}.
Fig.~\ref{fig:fisheye_erp}(d) displays a fisheye image projected onto an icosahedron at subdivision level 7.
Jiang et al. proposed UGSCNN, whereby images were projected onto an icosahedron mesh and ConvNets were applied as linear combinations of differential operators on the mesh with learnable parameters~\cite{jiang2019}. However, operators on the mesh structures are computationally less efficient compared to normal two-dimensional (2D) ConvNets on images. Therefore, 
Liu et al., Cohen et al., and Zhang et al. proposed unfolding and distorting the icosahedron grid so that normal 2D ConvNets could be applied on the icosahedron~\cite{liu2019, cohen2019, zhang2019}. Moreover, Zhang et al. proposed orientation-aware ConvNets for indoor scene semantic segmentation~\cite{zhang2019}.

In this study, we focus on all-around depth estimation in indoor environments from multiple omnidirectional images. 
We use the icosahedron to apply ConvNets to omnidirectional images to deal with the distortion.
Firstly, multiple omnidirectional images are projected onto the icosahedron and the image features are extracted by applying our proposed icosahedron-based ConvNets to the images on the icosahedron. Thereafter, a cost volume is generated from these features using our proposed icosahedron-based spherical sweeping and the cost volume is regularized using icosahedron-based 3D ConvNets. Finally, the depth is obtained by depth regression from the cost volume. Furthermore, we consider the orientation for the estimation as in~\cite{zhang2019}, because this is beneficial to estimating the depth in indoor environments and making our proposed method robust to camera alignments. Our proposed method is computationally efficient, so that the depth can be estimated from four fisheye images in less than a second using a laptop with a GPU. Therefore, it is suitable for real-world robotics applications.
 
In summary, our contributions are as follows:
\begin{enumerate}
\setlength{\parskip}{0pt}
    \item We propose a new method called IcoSweepNet for depth estimation from multiple fisheye images, which is robust to camera alignment by using the extrinsic camera parameters for the feature extraction.
    \item We propose a new icosahedron-based representation and computationally efficient ConvNets for omnidirectional images, which we name CrownConv because the representation resembles a crown made of origami.
    \item We propose icospherical sweeping that is icosahedron-based spherical sweeping to generate the cost volume from omnidirectional images.
\end{enumerate}

\begin{figure*}[tb]
    \centering
    \includegraphics[scale=0.82]{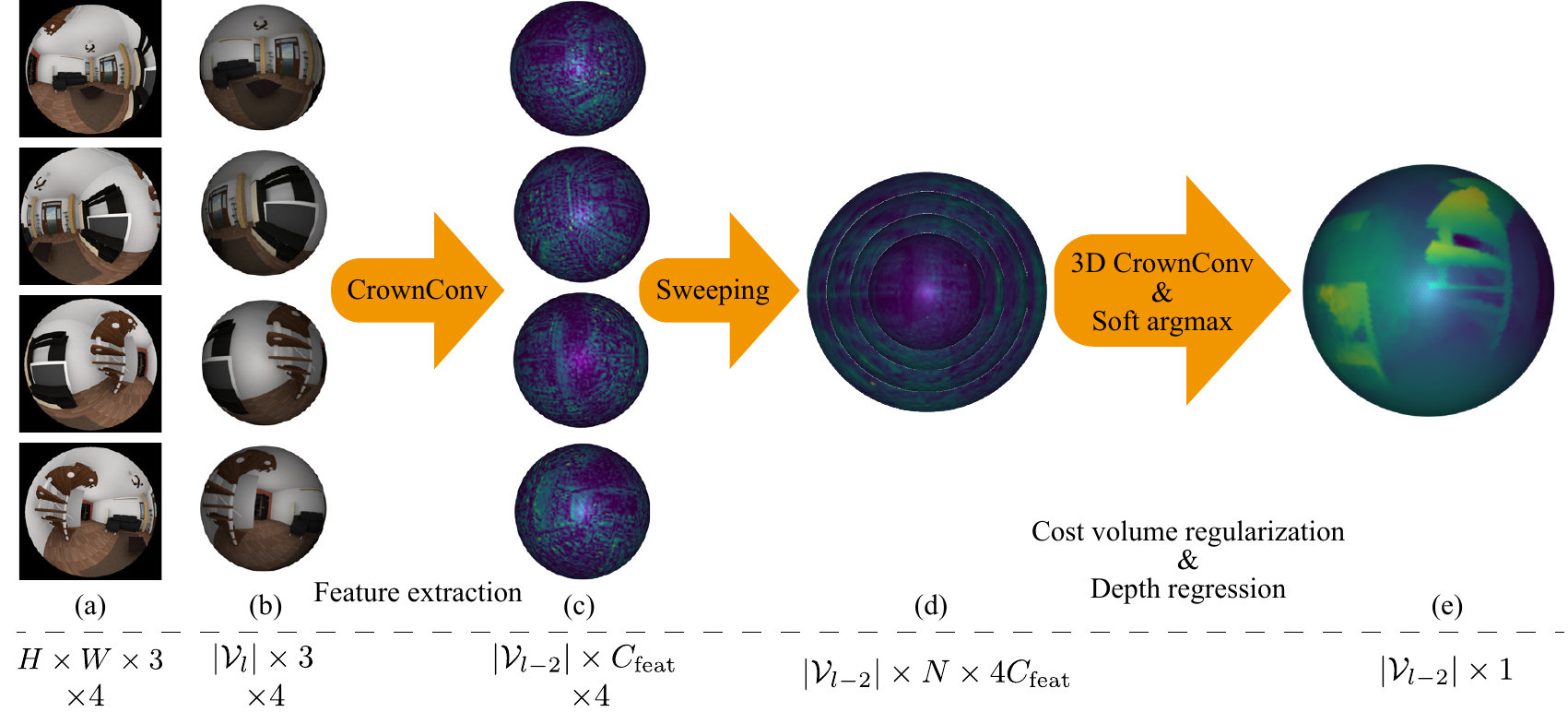}
    \caption{Overview of proposed method. (a) Original input of fisheye images. (b) Images projected onto icosahedron level $l$. (c) Image features extracted on icosahedron level $l-2$ using CrownConv. (d) Cost volume generated using icospherical sweeping. (e) Cost volume regularized using 3D CrownConv, with depth obtained by depth regression.
    The row below indicates the feature shapes.}
    \label{fig:overview}
\end{figure*}

\section{METHODS}
\subsection{Overview}
Our proposed method is similar to previous learning-based plane-sweeping methods. The greatest difference is that the operators are applied on 2D image planes in existing learning-based plane-sweeping methods, whereas the operators in our proposed method are all applied on the icosahedron. An overview is presented in Fig.~\ref{fig:overview}. 
Multiple fisheye cameras facing various directions are mounted so that the surrounding environments are captured.

Firstly, as can be observed in Figs.~\ref{fig:overview}(a) and (b), fisheye images are projected onto icosahedron level $l$ using the intrinsic and extrinsic camera parameters. The camera orientations are considered so that the image regions of the ceilings and floors are projected onto the north and south poles of the icosahedron, respectively, regardless of the camera alignments, as explained further in~\ref{subsec:proj_icosahedron}.

Secondly, as can be observed in Figs.~\ref{fig:overview}(b) and (c), features are extracted from images on the icosahedron by CrownConv, which is a ConvNet designed for features on an icosahedron. CrownConv and the feature extraction are explained in~\ref{subsec:crownconv} and~\ref{subsec:feature_extraction}, respectively.

Thirdly, as can be observed in Figs.~\ref{fig:overview}(c) and (d), a cost volume is generated from the features by icospherical sweeping, which is icosahedron-based spherical sweeping. Icospherical sweeping is explained in~\ref{subsec:icosweeping}.

Finally, as can be observed in Figs.~\ref{fig:overview}(d) and (e), the cost volume is regularized using 3D CrownConv, and the depth is obtained by depth regression, as discussed in~\ref{subsec:depth_regression}.

Prior to elaborating on our proposed method, we briefly explain the icosahedron in the following subsection.

\begin{figure}[tb]
    \centering
  \includegraphics[scale=0.75]{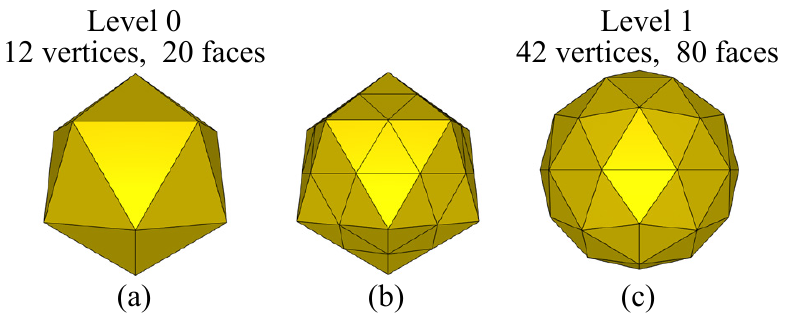}
    \caption{Subdivision process of icosahedron. (a) Icosahedron at level 0 with 12 vertices and 20 faces. (b) Dividing triangle face into four triangle faces. (c) Icosahedron at level 1 with 42 vertices and 80 faces by normalizing distance from center to each vertex following (b).}
      \label{fig:subdivision}
\end{figure}

\subsection{Icosahedron}
The regular icosahedron is one of the regular polyhedrons, which has 20 equilateral triangle faces and 12 vertices. The distance from the center to each vertex is the same, which is a suitable property for approximating the shape of a sphere. In this study, the distance is set to 1 and the center is set to the origin of the coordinate system; therefore, the icosahedron is suitable for approximation of the unit sphere. The resolution of the icosahedron can easily be increased by subdivision, whereby a triangle face is divided into four triangle faces. Figs.~\ref{fig:subdivision}(a), (b), and (c) present the subdivision of an icosahedron from level 0 (regular icosahedron) to level 1. The number of vertices and faces of the icosahedron at level $l$ are $2+10 \cdot 4^l$ and $20 \cdot 4^l$, respectively. In this case, $\mathcal{V}_l$, the set of vertices at level $l$, is formulated as follows:
\begin{equation}
\mathcal{V}_l =  \{\mathbf{v}^l_1, \mathbf{v}^l_2, \cdots, \mathbf{v}^l_{2+10 \cdot 4^l}\},
\end{equation}
where $\mathbf{v}^l_i \in \mathbb{R}^3 $ is a vertex of the icosahedron at level $l$ and $|\mathbf{v}^l_i|=1$ because it is normalized.

\begin{figure*}[tb]
    \centering
  \includegraphics[scale=0.9]{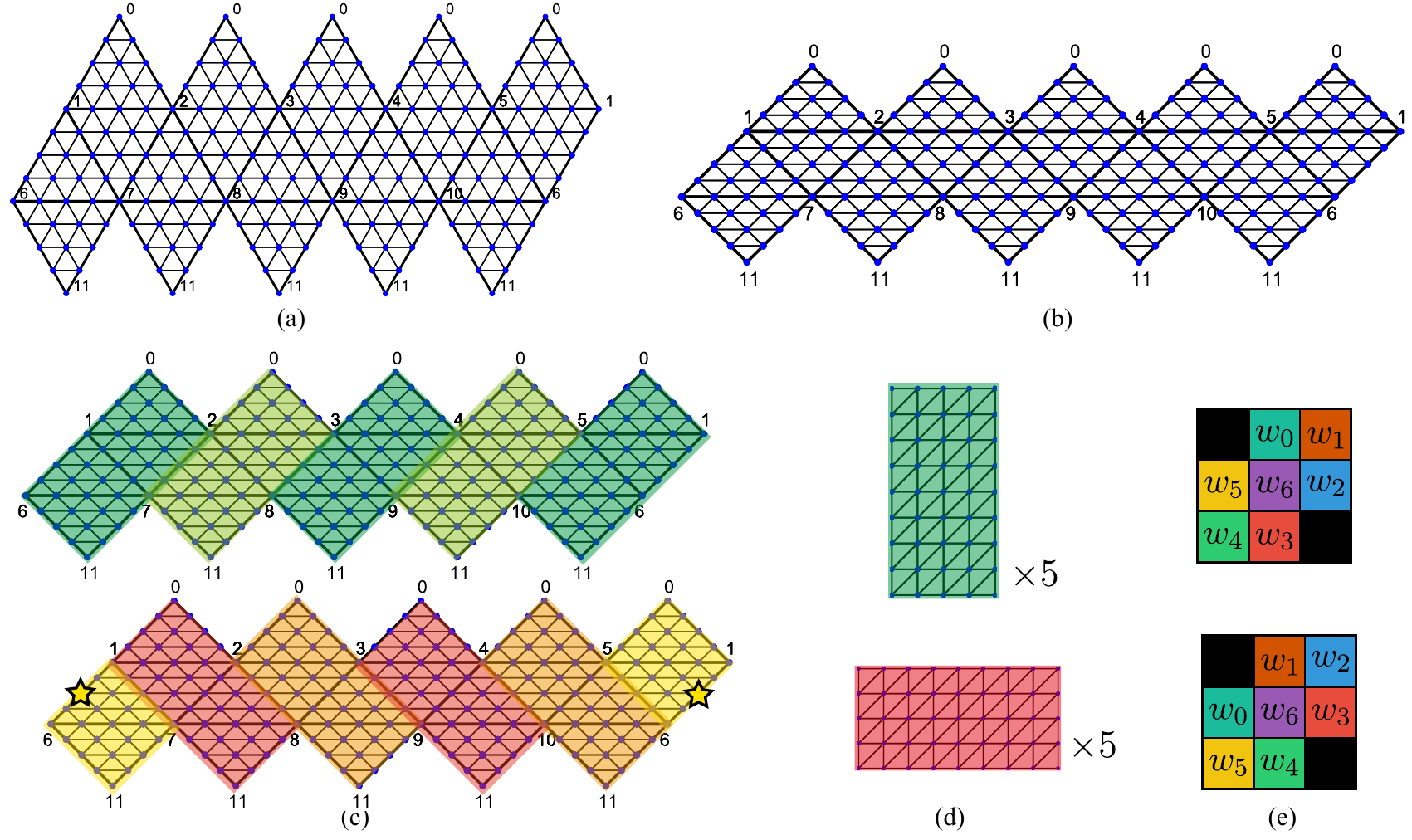}
    \caption{Origami crown representation for omnidirectional images. (a) Grid based on unfolded icosahedron at level 2. (b) Grid distorted so that normal ConvNet can be applied. 
    (c) Origami crown representation. Rectangles are extracted from the distorted grid in the two different directions. The stars at the ends of the grid indicate connectivity. (d) Final representation of five rectangles for two directions. (e) Weights of ConvNets shared but directions differ. The weights $w_0$ and $w_3$ are facing the longer direction. The numbers in (a), (b), and (c) represent the vertex indices of the icosahedron at level 0.}
      \label{fig:unfold_represen}
\end{figure*}
\begin{figure}[tb]
    \centering
  \includegraphics[scale=0.55]{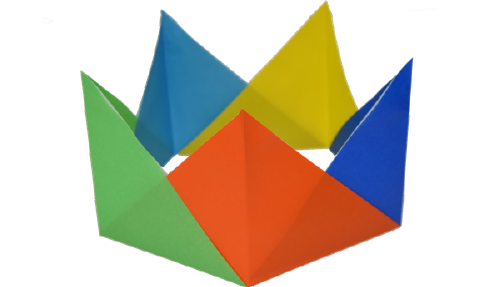}
    \caption{Crown made of origami.}
      \label{fig:origami_crown}
\end{figure}

\subsection{Projecting images onto icosahedron}
\label{subsec:proj_icosahedron}
The fisheye images are projected onto the icosahedron using the intrinsic and extrinsic camera parameters, which are estimated in advance.
We first define the intrinsic and extrinsic camera parameters.
The extrinsic parameter of camera $c_k$ is formulated as follows:
\begin{equation}
\mathbf{T}_{c_kw} = [\mathbf{R}_{c_kw}, \mathbf{t}_{c_kw}],
\end{equation}
where $\mathbf{R}_{c_kw} \in \text{SO}(3)$ and $\mathbf{t}_{c_kw} \in \mathbb{R}^3 $ are the rotation matrix and translation vector, respectively. Specifically, $\mathbf{T}_{c_kw}$ transforms a point $\mathbf{p}^w$ in the world coordinates to a point $\mathbf{p}^{c_k}$ in the camera $c_k$ coordinates. Moreover, $\mathbf{T}_{wc_k}$ is defined in the same manner, which transforms $\mathbf{p}^{c_k}$ into $\mathbf{p}^w$. The intrinsic parameter of $c_k$ is represented as a projection function $\pi_{c_k} \colon \mathbb{R}^3 \to \Omega$, which projects $\mathbf{p}^{c_k}$ onto a point $\mathbf{u}$ in the image domain $\Omega$. We use the fisheye models proposed in~\cite{scaramuzza2006, urban2015} for the projection. Our method can also be used on equirectangular images by replacing the fisheye projection models with equirectangular projection models.

We consider the direction of gravity, which is beneficial for estimating the depth in indoor environments simultaneously, which makes our proposed method robust to changes in the camera alignment. The motivation is that the image regions of the ceilings and floors will be projected onto the north and south poles of the icosahedron, respectively, regardless of the camera alignment. 
Therefore, a projected image on the icosahedron is formulated as follows:
\begin{equation}
^{\rm{ico}}\mathcal{I}^l_{c_k} = \{\mathcal{I}_{c_k}\left(\pi_{c_k} \left(\mathbf{R}_{c_kw} \mathbf{v}^l_i \right) \right) 
 \;  \mid \; \mathbf{v}^l_i \in \mathcal{V}_l \}, 
\label{eq:project_on_ico}
\end{equation}
where $^{\rm{ico}}\mathcal{I}^l_{c_k}$ is the projected image of the camera $c_k$ on the icosahedron at level $l$ 
and $\mathcal{I}_{c_k}(\mathbf{u})$ is the pixel value of the image of the camera $c_k$ at location $\mathbf{u}$.

\subsection{CrownConv}
\label{subsec:crownconv}
We first describe CrownConv, which is a ConvNet designed for features on an icosahedron. As in~\cite{liu2019, cohen2019, zhang2019}, we unfold and distort the icosahedron grid so that normal 2D ConvNets can be applied to the icosahedron. Figs.~\ref{fig:unfold_represen}(a) and (b) present the unfolded icosahedron grid and distorted unfolded icosahedron grid, respectively. 
Thereafter, we propose representing the features on the icosahedron as five rectangles in two different directions, namely upper right and lower right, as illustrated in Figs.~\ref{fig:unfold_represen}(c) and (d). 
As the representation in Fig.~\ref{fig:unfold_represen}(c) resembles a crown made of origami, we refer to it as the origami crown representation. An origami crown is presented in Fig.~\ref{fig:origami_crown} for reference. 
Subsequently, the features ($|\mathcal{V}_l| \times C $) on the icosahedron at level $l$ are converted into five vertical rectangle features ($\left[2^{l+1}+1 \right] \times \left[ 2^l+1\right] \times C $) and five horizontal rectangle features ($\left[ 2^l+1\right]  \times \left[2^{l+1}+1 \right] \times C $), as indicated in Fig.~\ref{fig:unfold_represen}(d), where $C$ is the number of channels of the features.
These are formulated as follows:
\begin{align}
\begin{split}
   ^{\rm{col}}\mathcal{G}^l, ^{\rm{row}}\mathcal{G}^l &=\Pi(\mathcal{F}^l),  \\
    ^{\rm{col}}\mathcal{G}^l &= \{^{\rm{col}}g^l_i  \;  \mid \; i \in 1,2,\cdots, 5 \}, \\
    ^{\rm{row}}\mathcal{G}^l &= \{^{\rm{row}}g^l_i  \;  \mid \; i \in 1,2,\cdots, 5 \},
\end{split}
\end{align}
where $\Pi(\cdot)$ is a function for the conversion and $\mathcal{F}^l$ is the features on the icosahedron at level $l$. $^{\rm{col}}\mathcal{G}^l$ and $^{\rm{row}}\mathcal{G}^l$ are the five vertical and horizontal rectangle features, respectively.

CrownConv is applied to $^{\rm{col}}\mathcal{G}^l$ and $^{\rm{row}}\mathcal{G}^l$ using normal 2D ConvNets. We use shared weights of the ConvNets but different directions for $^{\rm{col}}\mathcal{G}^l$ and $^{\rm{row}}\mathcal{G}^l$, similar to the process in~\cite{zhang2019}, as illustrated in Fig.~\ref{fig:unfold_represen}(e). The upper and lower sides of Fig.~\ref{fig:unfold_represen}(e) represent $^{\rm{col}}W$, which are the weights for $^{\rm{col}}\mathcal{G}^l$, and $^{\rm{row}}W$, which are the weights for $^{\rm{row}}\mathcal{G}^l$, respectively. As can be observed from Fig.~\ref{fig:unfold_represen}(e), the weights $w_0$ and $w_3$ face the longer directions. CrownConv is formulated as follows:
\begin{align}
\begin{split}
    ^{\rm{col}}\check{\mathcal{G}}^l &= \{ \opename{conv}(\opename{pad}(^{\rm{col}}g^l_i), ^{\rm{col}}W)   \;  \mid \; ^{\rm{col}}g^l_i \in ^{\rm{col}}\mathcal{G}^l \}, \\
    ^{\rm{row}}\check{\mathcal{G}}^l &= \{ \opename{conv}(\opename{pad}(^{\rm{row}}g^l_i), ^{\rm{row}}W)  \;  \mid \; ^{\rm{row}}g^l_i \in ^{\rm{row}}\mathcal{G}^l \}, 
    \label{eq:crown}
\end{split}
\end{align}
where $\check{x}$ means that CrownConv is applied to a feature $x$. Moreover, $\opename{conv}(x, W)$ and $\opename{pad}(\cdot)$ represent a convolution operator of $x$ with weight $W$ and a padding operator, respectively. 
Instead of zero padding, we apply padding with the replicated border to alleviate artifacts on the edges.
CrownConv is computationally efficient because all of the operators used in~\eqref{eq:crown} are the same for the 2D images, which are highly optimized for GPU computations.

After applying CrownConv, $^{\rm{col}}\check{\mathcal{G}}^l$ and $^{\rm{row}}\check{\mathcal{G}}^l$ should be integrated into the features on the icosahedron. 
Therefore, the inverse function of $\Pi(\cdot)$ is formulated as follows:
\begin{align}
\begin{split}
   \check{\mathcal{F}}^l &=\Pi^{-1}(^{\rm{col}}\check{\mathcal{G}}^l, ^{\rm{row}}\check{\mathcal{G}}^l ), \\
                 &= \{\frac{1}{|\check{\mathcal{G}}^{\mathbf{v}^l_i}|} \sum_{g \in \check{\mathcal{G}}^{\mathbf{v}^l_i}} g 
                 \;  \mid \; \mathbf{v}^l_i \in \mathcal{V}_l \},
\label{eq:integ}
\end{split}
\end{align}
where $\check{\mathcal{F}}^l$ represents the integrated features on the icosahedron and $\Pi^{-1}(\cdot)$ is the inverse function of $\Pi(\cdot)$. Furthermore,
$\check{\mathcal{G}}^{\mathbf{v}^l_i}$
is a set of values corresponding to $\mathbf{v}^l_i$ in $^{\rm{col}}\check{\mathcal{G}}^l$ and $^{\rm{row}}\check{\mathcal{G}}^l$.
We explain~\eqref{eq:integ} by means of an example: all 10 of the rectangle features have a value for $\mathbf{v}^l_0$, which is an index numbering zero in Fig.~\ref{fig:unfold_represen}(c). Then, the final value on the icosahedron is the mean of the values in the 10 rectangle features. This integration also enables the exchange of information between $^{\rm{col}}\check{\mathcal{G}}^l$ and $^{\rm{row}}\check{\mathcal{G}}^l$.

The concept of CrownConv is similar to HexConv~\cite{zhang2019}; therefore, the differences between CrownConv and HexConv are described here. For HexConv, five rectangles were extracted from the distorted unfolded icosahedron grid in only one direction, not two directions. Thereafter, HexConv copied features on the edges between five rectangles in every layer before applying normal 2D ConvNets, which helped HexConv extract global contextual information even if an object appeared across the line between vertices 2 and 7 in Fig.~\ref{fig:unfold_represen}(c). However, copying features in every layer prevented efficient computation. Meanwhile, CrownConv can extract global contextual information without copying features as the full shape of the object appears in either the five horizontal rectangles or the five vertical rectangles, which makes CrownConv more computationally efficient.

\subsection{Feature extraction}
\label{subsec:feature_extraction}
Our feature extraction module is based on ResNet~\cite{resnet}. 
Instead of normal 2D ConvNets, we use CrownConv to extract features from the images on the icosahedron. We place $\Pi^{-1}(\cdot)$ and $\Pi(\cdot)$ several times in the middle of the feature extraction module to exchange information between the rectangle features $^{\rm{col}}\check{\mathcal{G}}^l$ and $^{\rm{row}}\check{\mathcal{G}}^l$, and we place $\Pi^{-1}(\cdot)$ at the end of the module to obtain the extracted features on the icosahedron.
During the extraction, the features are downsampled twice by 
CrownConv with stride 2. 
Consequently, $\mathcal{F}^{l-2}_{c_k}$ is obtained from $^{\rm{ico}}\mathcal{I}^l_{c_k}$.
The shape of $\mathcal{F}^{l-2}_{c_k}$ is $|\mathcal{V}_{l-2}| \times C_{\rm{feat}}$, where $C_{\rm{feat}}$ is the number of channels of the extracted features.
For the detailed network architecture, see our GitHub page.

\begin{figure}[tb]
    \centering
  \includegraphics[scale=0.82]{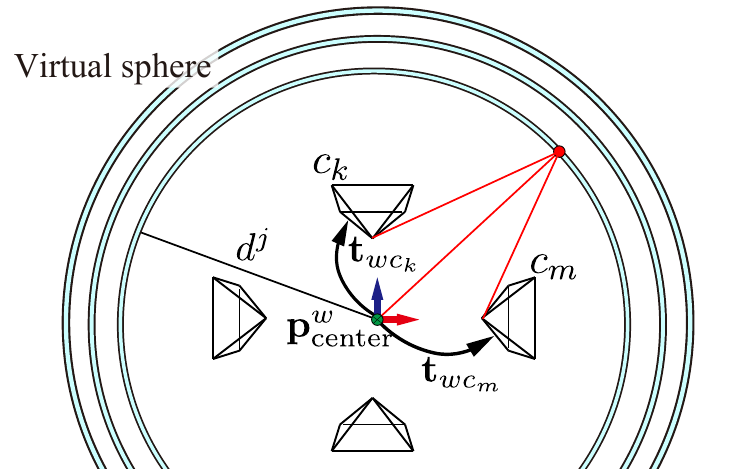}
    \caption{Illustration of icospherical sweeping. Virtual spheres are generated at several distances $d^j$ from the center of the camera rigs and the extracted features are projected onto the spheres to calculate the cost.}
      \label{fig:sweeping}
\end{figure}
\begin{figure}[tb]
    \centering
  \includegraphics[page=2,scale=0.75]{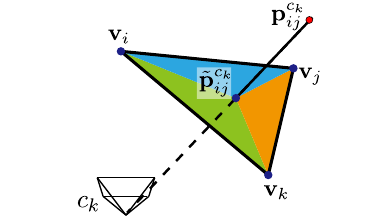}
    \caption{Triangle interpolation, where $\mathbf{p}^{c_k}_{ij}$ is projected onto the icosahedron of camera $c_k$ and the final value of the projected point $\tilde{\mathbf{p}}^{c_k}_{ij}$ is calculated by triangle interpolation of the neighboring vertex values.}
      \label{fig:triangle_interpolation}
\end{figure}
\subsection{Icospherical sweeping}
\label{subsec:icosweeping}
The extracted features $\mathcal{F}^{l-2}_{c_k}$ are warped by icospherical sweeping and a cost volume is generated on the icosahedron by concatenating the warped features. 
In icospherical sweeping, similar to spherical sweeping~\cite{im2016}, the features are projected onto the virtual spheres that are generated at several distances $d^j$ from the center of the camera rigs, as illustrated in Fig.~\ref{fig:sweeping}. 
The cost volume is formulated as follows:
\begin{align}
\begin{split}
V(i, j) &= \opename{concat} \left( \{ 
\opename{sample}\left(\mathcal{F}^{l-2}_{c_k}, \mathbf{p}^{c_k}_{ij} \right)
\;  \mid \;  c_k \in \mathcal{C} \} 
\right), \\
\mathbf{p}^{c_k}_{ij} &= \mathbf{p}^{w}_{\rm{center}} + d^j \cdot \mathbf{v}^{l-2}_i - \mathbf{t}_{wc_k}, \\
\end{split}
\label{eq:cost}
\end{align}
where $\opename{concat}(\cdot)$, $\mathcal{C}$, and $\mathbf{p}^{w}_{\rm{center}} \in \mathbb{R}^3$ are the concatenation operator, the set of all cameras, and the center of the camera rigs in the world coordinates, respectively.
$\mathbf{p}^{c_k}_{ij} \in \mathbb{R}^3$ is a point projected onto the virtual sphere in the camera $c_k$ coordinates. It should be noted that only the translation $\mathbf{t}_{wc_k}$ is required for the coordinate transformation, as the rotation has already been applied in~\eqref{eq:project_on_ico}.

Moreover, $\opename{sample}\left(\mathcal{F}^{l-2}_{c_k}, \mathbf{p}^{c_k}_{ij} \right)$ is the sampling operation by triangle interpolation, as illustrated in Fig.~\ref{fig:triangle_interpolation}. As can be observed in Fig.~\ref{fig:triangle_interpolation}, $\mathbf{p}^{c_k}_{ij}$ is projected onto the icosahedron of camera $c_k$ and the final value is calculated as follows:
\begin{align}
\begin{split}
\opename{sample}\left(\mathcal{F}^{l-2}_{c_k}, \mathbf{p}^{c_k}_{ij} \right) &=  
\mathcal{F}^{l-2}_{c_k}(\mathbf{v}_i) 
\frac{\mathcal{A}(\tilde{\mathbf{p}}^{c_k}_{ij}, \mathbf{v}_j, \mathbf{v}_k)}
{\mathcal{A}(\mathbf{v}_i, \mathbf{v}_j, \mathbf{v}_k)} \\
&+ \mathcal{F}^{l-2}_{c_k}(\mathbf{v}_j) 
\frac{\mathcal{A}(\mathbf{v}_i, \tilde{\mathbf{p}}^{c_k}_{ij}, \mathbf{v}_k)}
{\mathcal{A}(\mathbf{v}_i, \mathbf{v}_j, \mathbf{v}_k)} \\
&+ \mathcal{F}^{l-2}_{c_k}(\mathbf{v}_k) 
\frac{\mathcal{A}(\mathbf{v}_i, \mathbf{v}_j, \tilde{\mathbf{p}}^{c_k}_{ij})}
{\mathcal{A}(\mathbf{v}_i, \mathbf{v}_j, \mathbf{v}_k)},  \\
\tilde{\mathbf{p}}^{c_k}_{ij} &= \frac{\mathbf{p}^{c_k}_{ij}}{|\mathbf{p}^{c_k}_{ij}|},  
\end{split}
\label{eq:tri_inter}
\end{align}
where $\mathcal{A}(\mathbf{v}_i, \mathbf{v}_j, \mathbf{v}_k)$ represents the area of the triangle of vertices $\mathbf{v}_i$, $\mathbf{v}_j$, and $\mathbf{v}_k$, and $\mathcal{F}^{l-2}_{c_k}(\mathbf{v}_i)$ represents the value of $\mathcal{F}^{l-2}_{c_k}$ at vertex $\mathbf{v}_i$.
These operations depend on only the camera poses, depth from the center, and level of the icosahedron; therefore, the indices and factors of the interpolation are cached for efficient computation.

The distance $d^j$ from the center of the camera rigs is formulated as follows:
\begin{equation}
    \frac{1}{d^j} = \frac{j - 1}{N - 1} \cdot \frac{1}{d_{\rm{min}}} + \epsilon,  \quad j \in \{1,2,\dots,N \},
\end{equation}
where $N$, $d_{\rm{min}}$, and $\epsilon$ are the number of virtual spheres, the minimum distance of the virtual sphere, and a small positive number to avoid division by zero, respectively.

\subsection{Depth regression}
\label{subsec:depth_regression}
The cost volume $V$ is regularized using the 3D CrownConv to obtain $V^*$. The 3D CrownConv is defined in the same manner as CrownConv by adding another dimension.

The final depth is obtained as the inverse depth index on the icosahedron at level $l-2$ by means of depth regression from $V^*$.
We use soft argmax~\cite{gcnet} as the depth regression method to enable the inverse depth index to have a floating-point number.
The depth is calculated from the regularized cost volume $V^*$, as follows:
\begin{equation}
\hat{D}\left( i \right ) = \sum^{N}_{j=1}j \sigma\left(V^*\left(i, \cdot \right)\right)_j, \quad i \in \{1,2,\dots,|\mathcal{V}_{l-2}| \},
\label{eq:regression}
\end{equation}
where $\sigma(\cdot)$ is a softmax operation.

\subsection{Loss function}
\label{subsec:loss_function}

The loss function formulated below is used to minimize the errors between the prediction $\hat{D}$ and ground truth inverse depth index $D_{\rm{gt}}$:
\begin{equation}
L(\hat{D}, D_{\rm{gt}}) = 
 \frac{1}{|\mathcal{V}_{l-2}|} \sum_{i \in |\mathcal{V}_{l-2}|}  f_\delta \left(\hat{D}\left(i\right), D_{\rm{gt}}\left(i\right)\right),
\label{eq:loss}
\end{equation}
where $f_\delta$ is the Huber loss~\cite{huber} with $\delta = 1$ 
and $D_{\rm{gt}}$ is generated as follows:
\begin{equation}
D_{\rm{gt}}\left(i\right) = 1 + \frac{d_{\rm{min}}}{d_{\rm{gt}}\left(i\right)} \cdot \left(N - 1 \right),
\end{equation}
where $d_{\rm{gt}}$ is the ground truth depth map on the icosahedron, which is calculated by \eqref{eq:project_on_ico}.

\section{EXPERIMENTS}
\subsection{Datasets}
We used synthetic datasets, namely OmniThings and OmniHouse~\cite{won2019iccv}, for the training and evaluation. 
These datasets contain images from four fisheye cameras, depths from the center of the camera rig, and the extrinsic and intrinsic camera parameters. It should be noted that OmniThings and OmniHouse contain unique camera alignments; therefore, the extrinsic and intrinsic camera parameters are the same for both OmniThings and OmniHouse.

As the test sets of OmniThings and OmniHouse are not publicity available, we used the training sets in this experiment. We used 7,216 sets from OmniThings for training, 2,000 sets from OmniThings for validation, and 2,048 sets from OmniHouse for evaluation. The model with the lowest validation error during training was selected as the best model and evaluated. 

\begin{figure}[tb]
    \centering
  \includegraphics[scale=0.75]{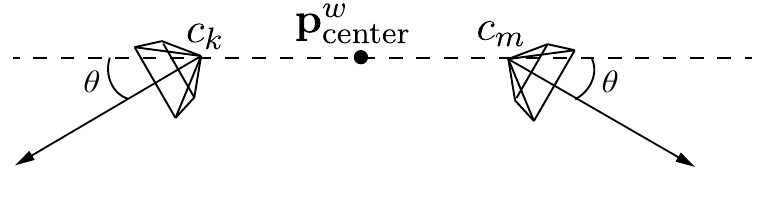}
    \caption{Camera alignment of rotated OmniHouse dataset. The cameras are facing downwards. Only two cameras are presented for simplicity. The arrows represent the camera-facing direction.}
      \label{fig:rotated_omnihouse}
\end{figure}
\subsection{Evaluation for robustness to camera alignment}
To evaluate the robustness to changes in the camera alignment, we used OmniHouse to create another dataset named as rotated OmniHouse, in which the camera-facing directions were changed.
In the original OmniHouse, the four cameras were facing the front, right, back, and left. Meanwhile, all cameras were facing downwards in rotated OmniHouse, as indicated in Fig.~\ref{fig:rotated_omnihouse}. We evaluated the proposed method on rotated OmniHouse at angles $\theta$ of 0\textdegree{} (the same as OmniHouse), 15\textdegree{}, 30\textdegree{}, and 45\textdegree{}. 
In the evaluation using rotated OmniHouse, certain regions of the estimated depth exhibited large errors owing to insufficient overlapping for stereo estimation. Therefore, those regions were excluded from the evaluation.

\subsection{Training details}
We used the PyTorch framework to implement the proposed network. The training was conducted in an end-to-end manner on four NVIDIA Tesla P100 GPUs with 16 GB of memory. 
We trained the network for 54,000 iterations with a batch size of 4. 
During training, Adam~\cite{adam} was used as the optimizer. The learning rate was set to 1e-3 for the first 36,000 iterations and 1e-4 for the remainder. 
The fisheye images were projected onto the icosahedron at level 7; thus, the depth was estimated on the icosahedron at level 5. Moreover, 
$d_{\rm{min}}$ was set to 0.55 m, whereas both $C_{\rm{feat}}$ and $N$ were set to 32. 

\subsection{Metrics}
The depth estimation results were evaluated using the same metrics as those in~\cite{won2019iccv}. The error was measured as follows:
\begin{equation}
E(i) = 100 \times \frac{ |\hat{D}(i) - D_{\rm{gt}}(i)|}{N}
\end{equation}
and the mean absolute error (MAE), root-mean-square error (RMS), and ratio (\%) of errors larger than $n$ (\textgreater $n$) were used as the evaluation metrics.

\section{RESULTS}
\subsection{Evaluation on rotated OmniHouse}
We compared our proposed method, referred to as IcoSweepNet, to OmniMVS~\cite{won2019iccv}. We implemented OmniMVS by ourselves\footnote{\url{https://github.com/matsuren/omnimvs_pytorch}} because the official implementation of OmniMVS is not publicity available at this time. We set the disparity number for OmniMVS as 48.

The evaluation results using rotated OmniHouse are presented in Table~\ref{tab:results}. As can be observed from Table~\ref{tab:results}, OmniMVS performed effectively when the camera alignment did not change substantially from the training. However, the performance of OmniMVS was degraded when the camera alignments were changed drastically from the training. Meanwhile, IcoSweepNet demonstrated stable performances with the different camera alignments. 

\begin{figure*}[tb]
    \centering
  \includegraphics[scale=0.8]{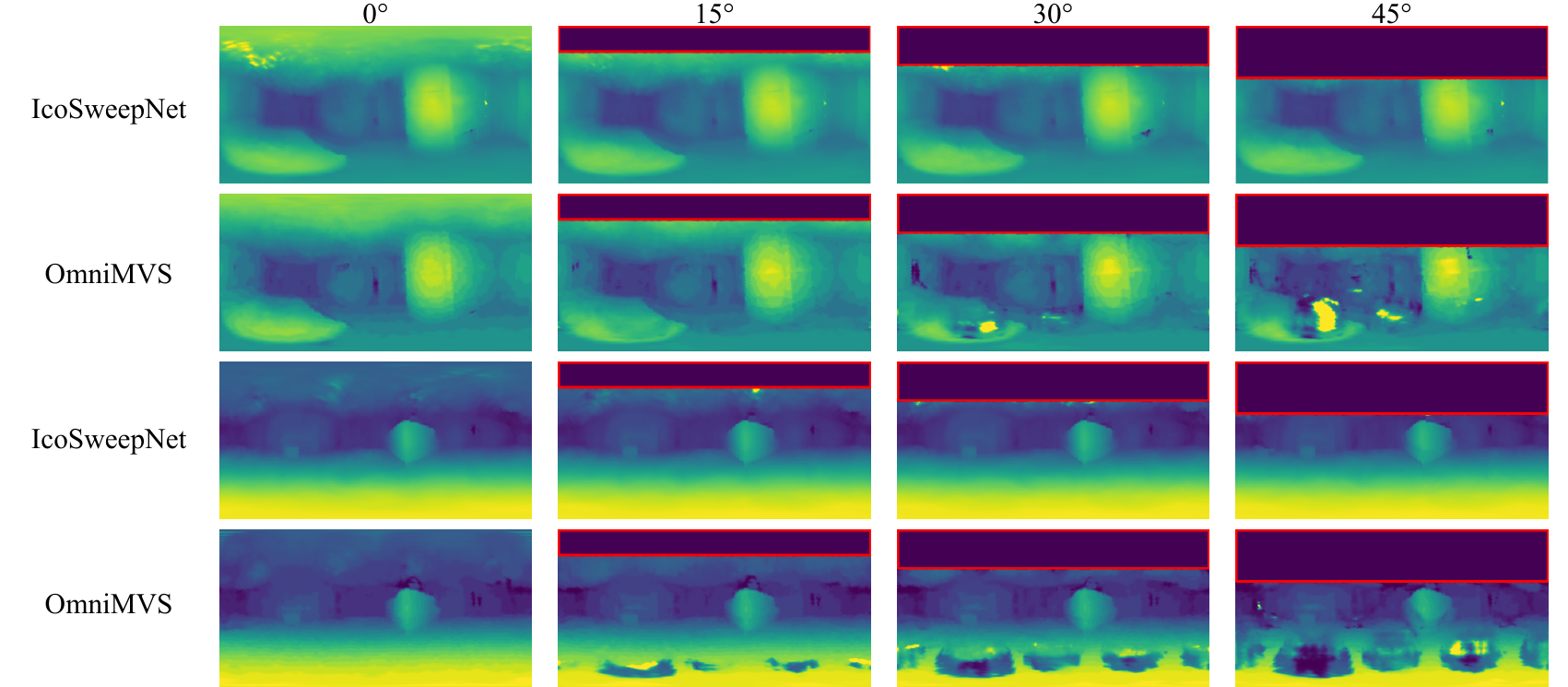}
    \caption{Examples of estimated depth maps. The depth estimations on the icosahedron are converted into equirectangular depth maps by linear interpolation for visualization. The model names are indicated on the left. The four columns from left to right correspond to the results on rotated OmniHouse at angles of 0\textdegree{}, 15\textdegree{}, 30\textdegree{}, and 45\textdegree{}. The red rectangle in each figure indicates that the region was masked out and excluded from the evaluation owing to insufficient overlapping for stereo estimation. }
      \label{fig:angle_results}
\end{figure*}

Examples of the estimated depth maps for rotated OmniHouse are presented in Fig.~\ref{fig:angle_results}. 
It can be observed from Fig.~\ref{fig:angle_results} that the depth maps estimated by OmniMVS contained more artifacts with more rotated datasets, whereas our model succeeded in estimating the depth maps despite the camera alignments being changed drastically.
\begin{table}
\caption{Evaluation results using rotated OmniHouse at different angles $\theta$. The best scores are indicated in bold.}
\centering
\setlength{\tabcolsep}{3pt}
\begin{tabular}{|cc|rrrrr|}
\hline
\multicolumn{2}{|c|}{} &  \multicolumn{5}{c|}{Error (smaller is better)} \\
Angle $\theta$ & Model & \textgreater 1 & \textgreater 3 & \textgreater 5 & MAE & RMS \\
\hline
\multirow{2}{*}{0\textdegree} & IcoSweepNet & 28.69 & 9.13 & 5.55 & 1.48 & 3.36 \\
 & OmniMVS & \bfseries{26.14} & \bfseries{5.33} & \bfseries{2.69} & \bfseries{1.05} & \bfseries{2.12} \\
\hline
\multirow{2}{*}{15\textdegree} & IcoSweepNet & \bfseries{27.46} & 8.39 & 4.93 & 1.34 & 3.05 \\
 & OmniMVS & 30.71 & \bfseries{6.78} & \bfseries{3.45} & \bfseries{1.24} & \bfseries{2.67} \\
\hline
\multirow{2}{*}{30\textdegree} & IcoSweepNet & \bfseries{26.86} & \bfseries{7.76} & \bfseries{4.48} & \bfseries{1.29} & \bfseries{2.99} \\
 & OmniMVS & 40.72 & 12.07 & 6.56 & 1.80 & 3.77 \\
\hline
\multirow{2}{*}{45\textdegree} & IcoSweepNet & \bfseries{26.30} & \bfseries{6.51} & \bfseries{3.60} & \bfseries{1.15} & \bfseries{2.53} \\
 & OmniMVS & 50.19 & 19.73 & 11.95 & 2.62 & 5.17 \\
\hline
\end{tabular}
\label{tab:results}
\end{table}

\subsection{Real-world experiment}
\begin{figure}[tb]
    \centering
  \includegraphics[scale=0.78]{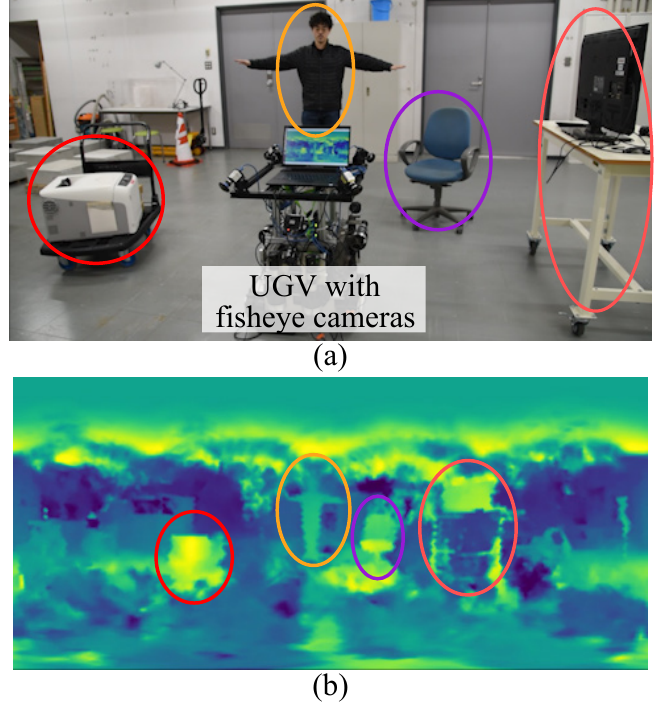}
    \caption{Example of real-world experiment. (a) Experimental scene. Four fisheye cameras were mounted on corners of a UGV. (b) Depth estimation visualized by equirectangular projection. The corresponding objects were circled with the same color. }
      \label{fig:realworld_result}
\end{figure}

Another experiment was conducted to demonstrate the performance in real world. 
We trained IcoSweepNet with $N$=16, 32 on a combination of OmniThings and OmniHouse datasets, and data augmentation with color jitting and random shift was applied to the fisheye images to mitigate the difference between synthetic data and real-world data.
Four fisheye cameras with FoV of 185{\textdegree} facing downward were mounted on corners of a UGV as illustrated in Fig.~\ref{fig:realworld_result}(a), and the depth was estimated from the four fisheye images using a laptop with NVIDIA GeForce RTX 2080 Max-Q.

IcoSweepNet estimated the depth from four fisheye images in 0.73 s ($N=32$) and 0.42 s ($N=16$) including capturing fisheye images, projecting the images on icosahedron, and estimating the depth. Therefore, our proposed method is applicable to robotics applications if the agility of the UGV is not very high.
An example of the estimated depth map is presented in Fig.~\ref{fig:realworld_result}(b). As can be observed in Fig.~\ref{fig:realworld_result}(b), the depths of the objects were estimated correctly, whereas the floor regions seemed to have wrong estimations. We believe it was because of reflections on the floor, which does not exist in synthetic datasets. Therefore, the training on real-world data may be suggested to improve the performance in real world.
The upper regions of Fig.~\ref{fig:realworld_result}(b) exhibited large errors owing to insufficient overlapping for stereo estimation.

\section{CONCLUSIONS}
We presented a method for estimating the all-round depth estimation from multiple omnidirectional images. We proposed a new origami crown representation of the icosahedron for omnidirectional images named as CrownConv, which is a ConvNet designed for the icosahedron.
Furthermore, icospherical sweeping was proposed for plane-sweeping stereo using omnidirectional images.
IcoSweepNet was demonstrated as robust to the camera alignments by using the extrinsic camera parameters; therefore, the proposed method performed effectively even when the camera alignment was changed drastically from the training. Finally, CrownConv is computationally efficient, and thus, our proposed method could estimate the depth from four fisheye images in less than a second using a laptop with a GPU. Therefore, it is suitable for real-world robotics applications.

Although only fisheye images were used in this study, the proposed method should be effective for equirectangular images without additional training, because all of the operators were applied on an icosahedron instead of 2D image planes.

In future work, improvements of the performance in real-world data and the inference in real-time are expected to make our model more applicable to robotics applications.


\addtolength{\textheight}{-0cm}   


\section*{ACKNOWLEDGMENT}
The authors would like to thank the members of the Intelligent Construction Systems Laboratory, The University of Tokyo for their useful suggestions, especially Mr. Shingo Yamamoto and Mr. Takumi Chiba from Fujita Corporation and Dr. Kazuhiro Chayama from KOKANKYO Engineering Corporation.

\end{document}